# Description Logics with Fuzzy Concrete Domains

**Straccia, Umberto**
ISTI - CNR, Pisa, ITALY
straccia@isti.cnr.it

## Abstract

We present a fuzzy version of description logics with concrete domains. Main features are: ($i$) concept constructors are based on t-norm, t-conorm, negation and implication; ($ii$) concrete domains are fuzzy sets; ($iii$) fuzzy modifiers are allowed; and ($iv$) the reasoning algorithm is based on a mixture of completion rules and bounded mixed integer programming.

## 1 INTRODUCTION

In the last decade a substantial amount of work has been carried out in the context of *Description Logics* (DLs) [1]. Nowadays, DLs have gained even more popularity due to their application in the context of the *Semantic Web* [7]. *Ontologies* play a key role in the Semantic Web. An ontology consists of a hierarchical description of important concepts in a particular domain, along with the description of the properties (of the instances) of each concept. Web content is then annotated by relying on the concepts defined in a specific domain ontology. DLs play a particular role in this context as they are essentially the theoretical counterpart of the *Web Ontology Language OWL DL*, a state of the art language to specify ontologies.

However, OWL DL becomes less suitable in domains in which the concepts to be represented have not a precise definition. As we have to deal with Web content, it is easily verified that this scenario is, unfortunately, likely the rule rather than an exception. For instance, just consider the case we would like to build an ontology about flowers. Then we may encounter the problem of representing concepts like "Candia is a creamy white rose with dark pink edges to the petals", "Jacaranda is a hot pink rose", "Calla is a very large, long white flower on thick stalks". As it becomes apparent such concepts hardly can be encoded into OWL DL, as they involve so-called *fuzzy* or *vague concepts*, like "creamy", "dark", "hot", "large" and "thick", for which a clear and precise definition is not possible.

The problem to deal *imprecision* has been addressed several decades ago by Zadeh ([20]), which gave birth in the meanwhile to the so-called *fuzzy set and fuzzy logic theory*. Unfortunately, despite the popularity of fuzzy set theory, relative little work has been carried out involving fuzzy DLs [5, 6, 10, 13, 15, 16, 18, 19].

Towards the management of vague concepts, we present a fuzzy extension of $\mathcal{ALC}(\mathbf{D})$ (the basic DL $\mathcal{ALC}$ [14] extended with concrete domains [9]). Main features are: ($i$) concept constructors are interpreted as t-norm, t-conorm, negation and implication. Current approaches consider conjunction as min, disjunction as max, negation as $1 - x$ only. Given the important role norm based connectives have in fuzzy logic, a generalization towards this directions is, thus, desirable; ($ii$) concrete domains are fuzzy sets. This has not been addressed yet in the literature and is a natural way to incorporate vague concepts with explicit membership functions into the language. This requirement has already been pointed out by Yen in [19], but not yet taken into account formally; ($iii$) fuzzy modifiers are allowed, similarly to [18, 6]; and ($iv$) reasoning is based on a mixture of completion rules and bounded Mixed Integer Programming (bMIP). The use of bMIP in our context is novel and allows for effective implementations. Fuzzy $\mathcal{ALC}(\mathbf{D})$ enhances current approaches to fuzzy DLs and is in line with [17], in which the need of a fuzzy extension of DLs in the context of the Semantic Web has been highlighted. In it, a fuzzy version of OWL DL has been presented without a calculus. Our work is a step forward in this direction, as it presents a calculus for an important sub-language of OWL DL. We also show that the computation is more complicated than the classical counterpart due to the generality of the connectives.

We proceed as follows. The following section presents fuzzy $\mathcal{ALC}(\mathbf{D})$. Section 3 presents the reasoning pro-cedure. Section 4 discusses related work, while Section 5 concludes and outlooks some topics for further research.

## 2 DESCRIPTION LOGICS WITH FUZZY DOMAINS

Fuzzy sets [20] allow to deal with vague concepts like `low pressure`, `high speed` and the like. A *fuzzy set A* with respect to a universe $X$ is characterized by a *membership function* $\mu_A: X \to [0, 1]$, or simply $A(x) \in [0, 1]$, assigning an $A$-membership degree, $A(x)$, to each element $x$ in $X$. $A(x)$ gives us an estimation of the belonging of $x$ to $A$. In fuzzy logics, the degree of membership $A(x)$ is regarded as the *degree of truth* of the statement "$x$ is $A$". Accordingly, in our fuzzy DL, a concept $C$ will be interpreted as a fuzzy set and, thus, concepts become *imprecise*; and, consequently, e.g. the statement "$a$ is an instance of concept $C$", will have a truth-value in $[0,1]$ given by the membership degree $C(a)$.

**Syntax.** Recall that $\mathcal{ALC}(\mathbf{D})$ is the basic DL $\mathcal{ALC}$ [14] extended with concrete domains [9] allowing to deal with data types such as strings and integers. In fuzzy $\mathcal{ALC}(\mathbf{D})$, however, concrete domains are fuzzy sets. A *fuzzy concrete domain* (or simply *fuzzy domain*) is a pair $\langle \Delta_\mathbf{D}, \Phi_\mathbf{D} \rangle$, where $\Delta_\mathbf{D}$ is an interpretation domain and $\Phi_\mathbf{D}$ is the set of *fuzzy domain predicates* $d$ with a predefined arity $n$ and an interpretation $d^\mathbf{D}: \Delta_\mathbf{D}^n \to [0, 1]$, which is a $n$-ary fuzzy relation over $\Delta_\mathbf{D}$. To the ease of presentation, we assume the fuzzy predicates have arity one, the domain is a subset of the rational numbers $\mathbb{Q}$ and the range is $[0,1] \cap \mathbb{Q}$ (in the following, whenever we write $[0,1]$, we mean $[0,1] \cap \mathbb{Q}$). For instance, we may define the predicate $\leq_{18}$ as an unary crisp predicate over the natural numbers denoting the set of integers smaller or equal to 18, i.e.

$$\leq_{18}(x) = \begin{cases} 1 & \text{if } x \leq 18 \\ 0 & \text{otherwise} \end{cases}.$$

On the other hand, `Young` may be a fuzzy domain predicate denoting the degree of youngness of a person's age with definition

$$\texttt{Young}(x) = \begin{cases} 1 & \text{if } x \leq 10 \\ (30-x)/20 & \text{if } 10 \leq x \leq 30 \\ 0 & \text{if } x \geq 30 \end{cases}.$$

Concerning fuzzy domain predicates, we recall that in fuzzy set theory and practice there are many membership functions for fuzzy sets membership specification. However, the *trapezoidal* $trz(x; a, b, c, d)$, the *triangular* $tri(x; a, b, c)$, the *L-function* (left shoulder function) $L(x; a, b)$ and the *R-function* (right shoulder function) $R(x; a, b)$ are simple, yet most frequently used to specify membership degrees (see Figure 1). Note that $tri(x; a, b, c) = trz(x; a, b, b, c)$. Also, we have that $\texttt{Young}(x) = L(x; 10, 30)$ holds.

We also consider fuzzy modifiers in fuzzy $\mathcal{ALC}(\mathbf{D})$. Fuzzy modifiers, like `very`, `more_or_less` and `slightly`, apply to fuzzy sets to change their membership function. Formally, a *modifier* is a function $f_m: [0,1] \to [0,1]$. For instance, we may define $\texttt{very}(x) = x^2$, while define $\texttt{slightly}(x) = \sqrt{x}$. Modifiers has been considered, for instance, in [6, 18].

Now, let C, $R_a$, $R_c$, $I_a$, $I_c$ and M be non-empty finite and pair-wise disjoint sets of *concepts names* (denoted $A$), *abstract roles names* (denoted $R$), *concrete roles names* (denoted $T$), *abstract individual names* (denoted $a$), *concrete individual names* (denoted $c$) and *modifiers* (denoted $m$). $R_a$ contains a non-empty subset $F_a$ of *abstract feature names* (denoted $r$), while $R_c$ contains a non-empty subset $F_c$ of *concrete feature names* (denoted $t$). Features are functional roles. The set of fuzzy $\mathcal{ALC}(\mathbf{D})$ *concepts* is defined by the following syntactic rules ($d$ is a unary fuzzy domain predicate):

$$\begin{aligned} C & \longrightarrow \top \mid \bot \mid A \mid C_1 \sqcap C_2 \mid C_1 \sqcup C_2 \mid \neg C \mid \\ & \quad \forall R.C \mid \exists R.C \mid \forall T.D \mid \exists T.D \mid m(C) \\ D & \longrightarrow d \mid \neg d \end{aligned}$$

A *TBox* $\mathcal{T}$ consists of a finite set of *terminological axioms* of the form $A \sqsubseteq C$ ($A$ is sub-concept of $C$) or $A = C$ ($A$ is defined as the concept $C$), where $A$ is a concept name and $C$ is concept. We also assume that no concept $A$ appears more than once on the left hand side of a terminological axiom and that no cyclic definitions are present in $\mathcal{T}$.[1] Note that in classical DLs, terminological axioms are of the form $C \sqsubseteq D$, where $C$ and $D$ are concepts. While from a semantics point of view it is easy to consider them as well (see [17]), we have not yet found a calculus to deal with such axioms.[2] Using axioms we may define the concept of a minor as

$$\texttt{Minor} = \texttt{Person} \sqcap \exists \texttt{age}. \leq_{18} \qquad (1)$$

while

$$\texttt{YoungPerson} = \texttt{Person} \sqcap \exists \texttt{age}.\texttt{Young} \qquad (2)$$

will denote a young person. Similarly, we may represent "Calla is a very large, long white flower on thick stalks" as `Calla = Flower ⊓`

---

[1] See [11].

[2] The problem relies on recursive definitions like $A \sqsubseteq \exists R.A$, which may generate an infinite computation in the model generation phase (assume that $A$ has an instance). For classical DLs, clever blocking conditions has been developed, which however do not exists yet for the fuzzy case.

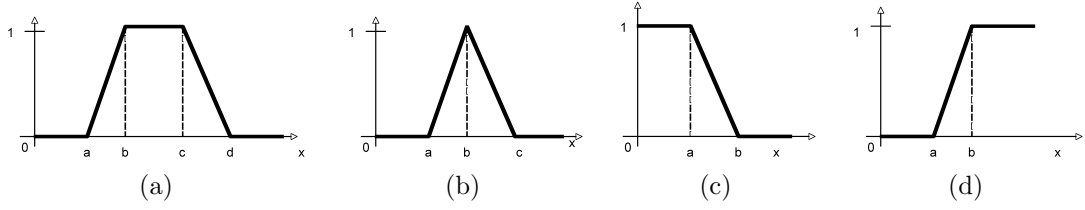

Figure 1: (a) Trapezoidal function; (b) Triangular function; (c) $L$-function; (d) $R$-function

($\exists$`hasSize.very`(`Large`)) $\sqcap$ ($\exists$`hasPetalWidth.Long`) $\sqcap$ ($\exists$`hasColour.White`) $\sqcap$ ($\exists$`hasStalks.Thick`), where `Large`, `Long` and `Thick` are fuzzy domain predicates and `very` is a concept modifier.

We also allow to formulate statements about individuals. A *concept-*, *role- assertion axiom* and an *individual (in)equality axiom* has the form $a\!:\!C$, $(a,b)\!:\!R$, $a \approx b$ and $a \not\approx b$, respectively, where $a,b$ are abstract individuals. For $n \in [0,1]$, an *ABox* $\mathcal{A}$ consists of a finite set of *fuzzy concept* and *fuzzy role assertion axioms* of the form $\langle \alpha, n \rangle$, where $\alpha$ is a concept or role assertion. Informally, $\langle \alpha, n \rangle$ constrains the truth degree of $\alpha$ to be greater or equal to $n$. Note that, like in [6, 15] one could add upper bounds to concept assertions, i.e. allow expressions of the form $\langle a\!:\!C \leq n \rangle$. To express upper bounds, we may use $\langle a\!:\!\ominus C, \ominus n \rangle$ instead. An ABox $\mathcal{A}$ may also contain a finite set of individual (in)equality axioms $a \approx b$ and $a \not\approx b$, respectively. A fuzzy $\mathcal{ALC}(\mathbf{D})$ *knowledge base* $\mathcal{K} = \langle \mathcal{T}, \mathcal{A} \rangle$ consists of a TBox $\mathcal{T}$ and an ABox $\mathcal{A}$.

**Semantics.** We generalize fuzzy $\mathcal{ALC}$ [15]. Unlike current approaches to fuzzy DLs, which deal with the interpretation of conjunction as min, disjunction as max, negation as $1-x$, our semantics of concept constructors is based on so-called *t-norm*, *t-conorm*, *negation* and *implication* [4]. So, let $\ominus, \otimes, \oplus$ and $\Rightarrow$ be a negation, a t-norm, a t-conorm and an implication function. Examples of functions are the following ($L$ stands for Lukasiewicz, $G$ stands for Gödel and $P$ for Product logic) . For negation: $\ominus_L x = 1 - x$, $\ominus_G 0 = 1$ and $\ominus_G x = 0$ if $x > 0$. For t-norms: $x \otimes_L y = \max(x+y-1, 0)$, $x \otimes_G y = \min(x,y)$, and $x \otimes_P y = x \cdot y$. For t-conorms: $x \oplus_L y = \min(x+y, 1)$, $x \oplus_G y = \max(x,y)$, and $x \oplus_P y = x + y - x \cdot y$. Concerning implication, we remind that it gives a truth-value to the formula $x \Rightarrow y$. Like for classical logic, we may use $x \Rightarrow y = \ominus x \oplus y$. For instance, $x \Rightarrow_{KD} y = \max(1-x, y)$ is the so-called Kleene-Dienes implication. Another approach to fuzzy implication is based on the so-called *residuum*. Its formulation is $x \Rightarrow y = \sup\{z \in [0,1]\!: x \otimes z \leq y\}$. Note that then $x \Rightarrow y = 1$ if $x \leq y$. If $x > y$ then, according to the chosen t-norm, we have that $x \Rightarrow_L y = 1 - x + y$, $x \Rightarrow_G y = y$ and $x \Rightarrow_P y = x/y$. Note also that $x \Rightarrow_L y = \ominus_L x \oplus_L y$. The same holds using Kleene-Dienes implication, Lukasiewicz negation and Gödel t-conorm. On the other hand $x \Rightarrow_P y \neq \ominus_G x \oplus_P y$. We conclude the discussion on fuzzy implication by noting that we have the following inferences: assume $x \geq n$ and $x \Rightarrow y \geq m$. Then $(i)$ under Kleene-Dienes implication we infer that if $n > 1-m$ then $y \geq m$ (this is used in [15]). $(ii)$ under residuum based implication w.r.t. a t-norm $\otimes$, we infer that $y \geq n \otimes m$, which we will use in this paper.

The semantics of fuzzy $\mathcal{ALC}(\mathbf{D})$ is as follows. A *fuzzy interpretation* $\mathcal{I}$ with respect to a concrete domain $\mathbf{D}$ is a pair $\mathcal{I} = (\Delta^{\mathcal{I}}, \cdot^{\mathcal{I}})$ consisting of a non empty set $\Delta^{\mathcal{I}}$ (called the *domain*), disjoint from $\Delta_{\mathbf{D}}$, and of a *fuzzy interpretation function* $\cdot^{\mathcal{I}}$ that assigns $(i)$ to each abstract concept $C \in \mathbf{C}$ a function $C^{\mathcal{I}}\!:\!\Delta^{\mathcal{I}} \to [0,1]$; $(ii)$ to each abstract role $R \in \mathtt{R}_a$ a function $R^{\mathcal{I}}\!:\!\Delta^{\mathcal{I}} \times \Delta^{\mathcal{I}} \to [0,1]$; $(iii)$ to each abstract feature $r \in \mathtt{F}_a$ a partial function $r^{\mathcal{I}}\!:\!\Delta^{\mathcal{I}} \times \Delta^{\mathcal{I}} \to [0,1]$ such that for all $u \in \Delta^{\mathcal{I}}$ there is an unique $w \in \Delta^{\mathcal{I}}$ on which $r^{\mathcal{I}}(u,w)$ is defined; $(iv)$ to each abstract individual $a \in \mathtt{I}_a$ an element in $\Delta^{\mathcal{I}}$; $(v)$ to each concrete individual $c \in \mathtt{I}_c$ an element in $\Delta_{\mathbf{D}}$; $(vi)$ to each concrete role $T \in \mathtt{R}_c$ a function $T^{\mathcal{I}}\!:\!\Delta^{\mathcal{I}} \times \Delta_{\mathbf{D}} \to [0,1]$; $(vii)$ to each concrete feature $t \in \mathtt{F}_c$ a partial function $t^{\mathcal{I}}\!:\!\Delta^{\mathcal{I}} \times \Delta_{\mathbf{D}} \to [0,1]$ such that for all $u \in \Delta^{\mathcal{I}}$ there is an unique $o \in \Delta_{\mathbf{D}}$ on which $t^{\mathcal{I}}(u,o)$ is defined; $(viii)$ to each modifier $m \in \mathtt{M}$ the function $f_m\!:\![0,1] \to [0,1]$; $(ix)$ to each unary concrete predicate $d$ the fuzzy relation $d^{\mathbf{D}}\!:\!\Delta_{\mathbf{D}} \to [0,1]$ and to $\neg d$ the negation of $d^{\mathbf{D}}$. The mapping $\cdot^{\mathcal{I}}$ is extended to concepts and roles as follows (where $u \in \Delta^{\mathcal{I}}$): $\top^{\mathcal{I}}(u) = 1$, $\bot^{\mathcal{I}}(u) = 0$,

$$
\begin{aligned}
(C_1 \sqcap C_2)^{\mathcal{I}}(u) &= C_1^{\mathcal{I}}(u) \otimes C_2^{\mathcal{I}}(u) \\
(C_1 \sqcup C_2)^{\mathcal{I}}(u) &= C_1^{\mathcal{I}}(u) \oplus C_2^{\mathcal{I}}(u) \\
(\neg C)^{\mathcal{I}}(u) &= \ominus C^{\mathcal{I}}(u) \\
(m(C))^{\mathcal{I}}(u) &= f_m(C^{\mathcal{I}}(u)) \\
(\forall R.C)^{\mathcal{I}}(u) &= \inf_{w \in \Delta^{\mathcal{I}}} R^{\mathcal{I}}(u,w) \Rightarrow C^{\mathcal{I}}(w) \\
(\exists R.C)^{\mathcal{I}}(u) &= \sup_{w \in \Delta^{\mathcal{I}}} R^{\mathcal{I}}(u,w) \otimes C^{\mathcal{I}}(w) \\
(\forall T.D)^{\mathcal{I}}(u) &= \inf_{o \in \Delta_{\mathbf{D}}} T^{\mathcal{I}}(u,o) \Rightarrow D^{\mathcal{I}}(o) \\
(\exists T.D)^{\mathcal{I}}(u) &= \sup_{o \in \Delta_{\mathbf{D}}} T^{\mathcal{I}}(u,o) \otimes D^{\mathcal{I}}(o) \, .
\end{aligned}
$$

The mapping $\cdot^{\mathcal{I}}$ is extended to assertion axioms as follows (where $a,b \in \mathtt{I}_a$): $(a\!:\!C)^{\mathcal{I}} = C^{\mathcal{I}}(a^{\mathcal{I}})$ and $((a,b)\!:\!R)^{\mathcal{I}} = R^{\mathcal{I}}(a^{\mathcal{I}}, b^{\mathcal{I}})$. The notion of *satisfiability* of a fuzzy axiom $E$ by a fuzzy interpretation $\mathcal{I}$, denoted $I \models E$, is defined as follows: $I \models A \sqsubseteq C$ iff for all $u \in \Delta^{\mathcal{I}}, A^{\mathcal{I}}(u) \leq C^{\mathcal{I}}(u)$ (this definition is equivalent to $[\inf_{u \in \Delta^{\mathcal{I}}} A^{\mathcal{I}}(u) \Rightarrow C^{\mathcal{I}}(u)] = 1$, which is derived directly from its FOL translation $\forall x. A(x) \Rightarrow C(x)$); $I \models A = C$ iff for all $u \in \Delta^{\mathcal{I}}, A^{\mathcal{I}}(u) = C^{\mathcal{I}}(u)$; $I \models \langle \alpha, n \rangle$ iff $\alpha^{\mathcal{I}} \geq n$; $\mathcal{I} \models a \approx b$ iff $a^{\mathcal{I}} = b^{\mathcal{I}}$; and $\mathcal{I} \models a \not\approx b$ iff $a^{\mathcal{I}} \neq b^{\mathcal{I}}$. The notion of *satisfiability* (is *model*) of a knowledge base $\mathcal{K} = \langle \mathcal{T}, \mathcal{A} \rangle$ and *entailment* of an assertional axiom is straightforward. Concerning terminological axioms, we also introduce degrees of subsumption. We say that $\mathcal{K}$ entails $A \sqsubseteq B$ to degree $n \in [0,1]$, denoted $\mathcal{K} \models \langle A \sqsubseteq B, n \rangle$ iff for every model $\mathcal{I}$ of $\mathcal{K}$, $[\inf_{u \in \Delta^{\mathcal{I}}} A^{\mathcal{I}}(u) \Rightarrow B^{\mathcal{I}}(u)] \geq n$.

**Example 1** *Consider the following simplified excerpt of a knowledge base about cars:*

$$\texttt{SportsCar} = \exists\texttt{speed}.\texttt{very}(\texttt{High}),$$
$$\langle \texttt{mg\_mgb}: \exists\texttt{speed}.\leq_{170}, 1 \rangle$$
$$\langle \texttt{ferrari\_enzo}: \exists\texttt{speed}.>_{350}, 1 \rangle,$$
$$\langle \texttt{audi\_tt}: \exists\texttt{speed}.=_{243}, 1 \rangle$$

speed *is a concrete feature. The fuzzy domain predicate* High *has membership function* $\texttt{High}(x) = R(x; 80, 250)$. *It can be shown that*

$$\mathcal{K} \models \langle \texttt{mg\_mgb}: \neg\texttt{SportsCar}, 0.72 \rangle$$
$$\mathcal{K} \models \langle \texttt{ferrari\_enzo}: \texttt{SportsCar}, 1 \rangle$$
$$\mathcal{K} \models \langle \texttt{audi\_tt}: \texttt{SportsCar}, 0.92 \rangle \ .$$

*Note how the maximal speed limit of the* mg_mgb *car ($\leq 170$) induces an upper limit, $0.28 = 1 - 0.72$, on the membership degree of being* mg_mgb *a* SportsCar.

**Example 2** *Consider $\mathcal{K}$ with terminological axioms (1) and (2). Then under Zadeh logic $\mathcal{K} \models \langle \texttt{Minor} \sqsubseteq \texttt{YoungPerson}, 0.5 \rangle$ holds (see Example 3).*

Finally, given $\mathcal{K}$ and an axiom $\alpha$, it is of interest to compute its best lower degree bound. The *greatest lower bound* of $\alpha$ w.r.t. $\mathcal{K}$, denoted $glb(\mathcal{K}, \alpha)$, is $glb(\mathcal{K}, \alpha) = \sup\{n: \mathcal{K} \models \langle \alpha, n \rangle\}$, where $\sup \emptyset = 0$. Determining the *glb* is called the *Best Degree Bound* (BDB) problem. For instance, the entailments in Examples 1 and 2 are the best possible degree bounds. Note that, $\mathcal{K} \models \langle \alpha, n \rangle$ iff $glb(\mathcal{K}, \alpha) \geq n$. Therefore, the BDB problem is the major problem we have to consider in fuzzy $\mathcal{ALC}(\mathbf{D})$, which we address in the next section.

We finally point out that the expressions of a knowledge base should not necessarily be considered as the language to be presented to the user, but rather are the *internal* representation of a reasoning system. Indeed, it may be questionable whether a user may be able or should be allowed to express $\langle \alpha, n \rangle$ as she might be unsure which value to chose for $n$. In case a user is not allowd to specify a truth value, alternative options might be: assume that $\mathcal{K} = \langle \mathcal{T}, \mathcal{A} \rangle$ consists of a TBox $\mathcal{T}$, but where the ABox $\mathcal{A}$ is a set of (unweighted) assertions $\alpha$ only. For $n \in [0, 1]$, let $\langle \mathcal{K}, n \rangle$ be the fuzzy knowledge base $\langle \mathcal{T}, \langle \mathcal{A}, n \rangle \rangle$, where $\langle \mathcal{A}, n \rangle$ is the ABox of fuzzy assertions $\{\langle \alpha, n \rangle: \alpha \in \mathcal{A}\}$. Then alternative definitions of the BDB problem are, for instance, (i) $glb(\mathcal{K}, \alpha) = \sup\{n: \langle \mathcal{K}, 1 \rangle \models \langle \alpha, n \rangle\}$; and (ii) $glb(\mathcal{K}, \alpha) = \sup\{n: \langle \mathcal{K}, n \rangle \models \langle \alpha, n \rangle\}$. Under the assumptions described below, these problems can still be solved by the method presented in the next section.

## 3 REASONING IN FUZZY $\mathcal{ALC}(\mathbf{D})$

To make our proof system for fuzzy $\mathcal{ALC}(\mathbf{D})$ correct and complete, we will assume that the chosen t-norm $\otimes$, t-conorm $\oplus$, negation $\ominus$ and implication $\Rightarrow$ are such that always $x \oplus y \equiv \ominus(\ominus x \otimes \ominus y)$; $x \Rightarrow y \equiv \ominus x \oplus y$; and $\ominus \forall x. A(x) \equiv \exists x. \ominus A(x)$ hold for all fuzzy sets $A$, where $\forall$ is interpreted as inf and $\exists$ as sup. These are true, e.g. for Lukasiewicz logic, but not for Gödel logic. [3]

Due to the restrictions on the chosen fuzzy functions, we do have that $(\forall R.C)^{\mathcal{I}} = (\neg \exists R. \neg C)^{\mathcal{I}}$. This will allow us to transform concept expressions into a semantically equivalent *Negation Normal Form* (NNF), *which is obtained by pushing in the usual manner negation on front of concept names, modifiers and concrete predicate names only*. With $\texttt{nnf}(C)$ we denote the NNF of concept $C$.

Additionally, we assume the set of truth degrees in $[0, 1]$ we will deal with is finite. From a practical point of view this is a limitation we can live with, especially taking into account that computers have finite resources, and thus, only a finite set of truth degrees can be represented. In particular, this includes our case were we use the rational numbers in $[0, 1] \cap \mathbb{Q}$ under a given fixed precision $p$ of numbers a computer can work with.

The basic idea behind our reasoning algorithm is as follows. Consider $\mathcal{K} = \langle \mathcal{T}, \mathcal{A} \rangle$. In order to solve the BDB problem, we combine appropriate DL completion rules with methods developed in the context of *Many-Valued Logics* (MVLs) [3]. In order to determine e.g. $glb(\mathcal{K}, a:C)$, we consider an expression of the form $\langle a: \neg C, \ominus x \rangle$ (informally, $\langle a: C \leq x \rangle$), where $x$ is a $[0, 1]$-valued variable. Then we construct a tableaux for $\mathcal{K} = \langle \mathcal{T}, \mathcal{A} \cup \{\langle a: \neg C, \ominus x \rangle\} \rangle$ in which the application of satisfiability preserving rules generates new assertion axioms together with *inequations* over $[0, 1]$-valued variables. These inequations have to hold in order to respect the semantics of the DL constructors. Finally, in order to determine the greatest lower bound, we *minimize* the original variable $x$ such that all constraints are satisfied [4]. In general, depending on the semantics of the DL constructors and fuzzy domain predicates we may end up with a general, bounded *Non Linear Programming* optimization problem. In this paper, however, we will limit the choice of the semantics of concept constructors, modifiers and fuzzy domain predicates in such a way that we end up with a *bounded Mixed Integer Program* (bMIP) optimization problem [12]. Interestingly, as for the MVL case, the tableaux we are generating contains *one* branch only and, thus, just *one* bMIP problem has to be solved.

---
[3] It is worth noting that (see [5]) the axiom $\top \sqsubseteq \neg(\forall R.A) \sqcap (\neg \exists R. \neg A)$ has no classical model. However, in [5] it is shown that in Gödel logic it has no finite model, but has an infinite model.

**Mixed Integer Programming.** A general MIP problem consists in minimizing a linear function with respect to a set of constraints that are linear inequations in which rational and integer variables can occur. In our case, the variables are bounded. More precisely, let $\mathbf{x} = \langle x_1, \ldots, x_k \rangle$ and $\mathbf{y} = \langle y_1, \ldots, y_m \rangle$ be variables over $\mathbb{Q}$, over the integers and let $A, B$ be integer matrices and $h$ an integer vector. The variables in $\mathbf{y}$ are called *control variables*. Let $f(\mathbf{x}, \mathbf{y})$ be an $k + m$-ary linear function. Then the *general MIP problem* is to find $\bar{\mathbf{x}} \in \mathbb{Q}^k, \bar{\mathbf{y}} \in \mathbb{Z}^m$ such that $f(\bar{\mathbf{x}}, \bar{\mathbf{y}}) = \min\{f(\mathbf{x}, \mathbf{y}) : A\mathbf{x} + B\mathbf{y} \geq h\}$. The general case can be restricted to what concerns the paper as we can deal with *bounded* MIP (bMIP). That is, the rational variables range over $[0, 1]$, while the integer variables ranges over $\{0, 1\}$. It is well known that the bMIP problem is NP-complete (for the belonging to NP, guess the $\mathbf{y}$ and solve in polynomial time the linear system, NP-hardness follows from NP-Hardness of 0-1 Integer Programming). Furthermore, we say that $M \subseteq [0, 1]^k$ is *bMIP-representable* iff there is a bMIP $(A, B, h)$ with $k$ real and $m$ 0-1 variables such that $M = \{\mathbf{x} : \exists \mathbf{y} \in \{0, 1\}^m$ such that $A\mathbf{x} + B\mathbf{y} \geq h\}$. In general, we require that a constructor $f$ is bMIP representable. In particular, the sets $g(f) = \{\langle x_1, \ldots, x_k, x \rangle : f(x_1, \ldots, x_k) \geq x\}$ and $\bar{g}(f) = \{\langle x_1, \ldots, x_k, x \rangle : f(x_1, \ldots, x_k) \leq x\}$ should be bMIP-representable. Interestingly, once a bMIB representation of a constructor is given, then sound, complete and linear tableaux rules can be obtained from it. Also, using ideas from *disjunctive programming*, the tableaux rules can be designed in such a way that a one-branch tree only is generated. See [3] for more on this issue and on bMIP-representabilty conditions for connectives. For instance, classical logic, Zadeh's fuzzy logic, and Lukasiewicz connectives, are bMIP-representable, while Gödel negation is not. In general, connectives whose graph can be represented as the union of a finite number of convex polyhedra are bMIB-representable [8], however, discontinuous functions may not be bMIP representable.

**The BDB problem.** We start with some preprocessing steps as for classical DLs [11]. First, each terminological axiom $A \sqsubseteq C \in \mathcal{T}$ can be replaced with $A = C \sqcap A^*$, where $A^*$ is a new concept name. Let $\mathcal{K}'$ the obtained knowledge base. Second, the newly obtained $\mathcal{K}'$ can be *expanded* by substituting every concept name $A$ occurring in $\mathcal{K}$, which is defined in $\mathcal{T}$, with its defining term in $\mathcal{T}$. Although, the expanded knowledge base may become of exponential size, the properties from a semantics point of view are left unchanged. Let $\mathcal{K}''$ the obtained knowledge base. Finally, each concept occurring in $\mathcal{K}''$ is then transformed into NNF. This last operations does not affect the semantics due to the restrictions we made on the fuzzy constructors. Notice that negation may appear on front of modifiers in the from $\neg m(C)$, where $C$ is a complex concept. Now, let V be a new alphabet of variables $x$ ranging over $[0, 1]$, W be a new alphabet of 0-1 variables $y$. We extend fuzzy assertions to the form $\langle \alpha, l \rangle$, where $l$ is a linear expression over variables in V, W and real values. A *linear constraint* is of the form $l \geq l'$ or $l \leq l'$, where $l, l'$ are linear expressions over variables in V, W and rational values. The satisfiability notion of linear constraints is immediate. A *constraint set* $S$ is a set of terminological axioms, fuzzy assertion axioms, (in)equality axioms and linear constraints. $\mathcal{I}$ *satisfies* $S$ iff $\mathcal{I}$ satisfies all elements of it. With $S_0$ we denote the constraint set $S_0 = \mathcal{T} \cup \mathcal{A}$. We will see later how to determine the satisfiability of a constraint set.

In the following, we assume that $S_0$ is satisfiable, otherwise $glb(\mathcal{K}, \alpha) = 1$. Note that our algorithm can be used to test the satisfiability of $S_0$ in the first place. As in [15], concerning fuzzy role assertions, we have that $\mathcal{K} \models \langle (a, b) : R, n \rangle$ iff $\langle (a, b) : R, m \rangle \in \mathcal{A}$ with $m \geq n$. Therefore, $glb(\mathcal{K}, (a, b) : R)) = \max\{n : \langle R(a, b), n \rangle \in \mathcal{A}\}$. So we do not consider this case further. Now, let us determine $glb(\mathcal{K}, a : C)$. As anticipated, $glb(\mathcal{K}, a : C)$ is determined by the minimal value of $x$ such that the constraint set $S = S_0 \cup \{\langle a : \neg C, \ominus x \rangle\}$ is satisfiable. Similarly, for a terminological axiom $A \sqsubseteq B$, we can compute $glb(\mathcal{K}, A \sqsubseteq B)$ as the minimal value of $x$ such that the constraint set $S = S_0 \cup \{\langle a : A \sqcap \neg B, \ominus x \rangle\}$ is satisfiable, where $a$ is new abstract individual. Therefore, the BDB problem can be reduced to minimal satisfiability problem.

Note that we previously gave also alternative definitions of the BDB problem. These cases can be reduced to the satisfiability problem as well. Indeed, (i) for $glb(\mathcal{K}, a : C) = \sup\{n : \langle \mathcal{K}, 1 \rangle \models \langle a : C, n \rangle\}$, deter-mine the minimal value of $x$ such that the constraint set $S = \mathcal{T} \cup \langle \mathcal{A}, 1 \rangle \cup \{\langle a : \neg C, \ominus x \rangle\}$ is satisfiable; while (ii) for $glb(\mathcal{K}, a : C) = \sup\{n : \langle \mathcal{K}, n \rangle \models \langle a : C, n \rangle\}$, determine the minimal value of $x$ such that the constraint set $S = \mathcal{T} \cup \langle \mathcal{A}, x \rangle \cup \{\langle a : \neg C, \ominus x \rangle\}$ is satisfiable. We will deserve to this issue more space in the extended version of this paper.

---
[4]Informally, suppose the minimal value is $\bar{n}$. We will know then that for any interpretation $\mathcal{I}$ satisfying the knowledge base such that $(a : C)^{\mathcal{I}} < \bar{n}$, the starting set is unsatisfiable and, thus, $(a : C)^{\mathcal{I}} \geq \bar{n}$ has to hold. Which means that $glb(\mathcal{K}, (a : C)) = \bar{n}$

**The Satisfiability problem.** We assume that the concept constructors, concept modifiers and fuzzy domains predicates are bMIB representable (as e.g., the membership functions in Figure 1). In particular, we present a correct and complete proof system where the DL connectives are interpreted according to Zadeh logic, while modifiers and fuzzy domain predicates are specified as a combination of linear functions over $[0, 1]$ and $\mathbb{Q}$, respectively, *as specified in Appendix A*. We also present a correct and complete proof system for Luaksiewicz logic in Appendix B.

Our satisfiability checking calculus is based on a set of constraint propagation rules transforming a set $S$ of constraints into "simpler" satisfiability preserving constraint sets $S_i$ until either $S_i$ contains a *clash* or no rule can be further be applied to $S_i$. If $S_i$ contains a clash then $S_i$ and, thus $S$ is immediately not satisfiable. Otherwise, we apply a bMIP oracle to solve the set of linear constraints in $S_i$ to determine either the satisfiability of the set or the minimal value for a given variable $x$, making $S_i$ satisfiable. We assume that a constraint set $S$ is reflexive, symmetric and transitively closed concerning the equality axioms. $S$ contains a *clash* iff either $\langle a : \bot, n \rangle \in S$ with $n > 0$, or $\{a \approx b, a \not\approx b\} \subseteq S$. The rules follow easily from the bMIP representations. *Each rule instantiation is applied at most once.* Before we can formulate the rules we need a technical definition involving feature roles (see [9]). Let $S$ be a constraint set, $r$ an abstract feature and both $\langle (a, b_1) : r, l_1 \rangle$ and $\langle (a, b_2) : r, l_2 \rangle$ occur in $S$. Then we call such a pair a *fork*. As $r$ is a function, such a fork means that $b_1$ and $b_2$ have to be interpreted as the same individual. A fork $\langle (a, b_1) : r, l_1 \rangle$, $\langle (a, b_2) : r, l_2 \rangle$ can be deleted by replacing all occurrences of $b_2$ in $S$ by $b_1$. A similar argument applies to concrete feature roles. At the beginning, we remove the forks from $S_0$. We assume that forks are eliminated as soon as they appear (as part of a rule application) with the proviso that newly generated individuals are replaced by older ones and not vice-versa. With $x_\alpha$ we denote the variable associated to the *atomic assertion* $\alpha$ of the form $a : A$ or $(a, b) : R$. $x_\alpha$ will take the truth value associated to $\alpha$, while with $x_c$ we denote the variable associated to the concrete individual $c$. The rules are the following:

**R**$A$. If $\langle \alpha, l \rangle \in S_i$ and $\alpha$ is an atomic assertion of the form $a : A$ or $(a, b) : R$ then $S_{i+1} = S_i \cup \{x_\alpha \geq l\}$.

**R**$\bar{A}$. If $\langle a : \neg A, l \rangle \in S_i$ then $S_{i+1} = S_i \cup \{x_{a : A} \leq 1 - l\}$.

**R**$\sqcap$. If $\langle a : C \sqcap D, l \rangle \in S_i$ then $S_{i+1} = S_i \cup \{\langle a : C, l \rangle, \langle a : D, l \rangle\}$.

**R**$\sqcup$. If $\langle a : C \sqcup D, l \rangle \in S_i$ then $S_{i+1} = S_i \cup \{\langle a : C, x_1 \rangle, \langle a : D, x_2 \rangle, x_1 + x_2 = l, x_1 \leq y, x_2 \leq 1 - y, x_i \in [0, 1], y \in \{0, 1\}\}$, where $x_i$ is a new variable, $y$ is a new control variable.

**R**$\exists$. If $\langle a : \exists R.C, l \rangle \in S_i$ then $S_{i+1} = S_i \cup \{\langle (a, b) : R, l \rangle, \langle b : C, l \rangle\}$, where $b$ is a new abstract individual. The case for concrete roles is similar.

**R**$\forall$. If $\{\langle a : \forall R.C, l_1 \rangle, \langle (a, b) : R, l_2 \rangle\} \subseteq S_i$ then $S_{i+1} = S_i \cup \{\langle a : C, x \rangle, x + y \geq l_1, x \leq 1 - y, l_1 + l_2 \leq 2 - y, x \in [0, 1], y \in \{0, 1\}\}$, where $x$ is a new variable and $y$ is anew control variable. The case for concrete roles is similar.

**R**$m$. If $\langle a : m(C), l \rangle \in S_i$ then $S_{i+1} = S_i \cup \gamma(a : C, l)$, where the set $\gamma(a : C, l)$ is obtained from the bMIP representation (see appendix) of $g(m)$ as follows: replace in $g(m)$ all occurrences of $x_2$ with $l$. Then resolve for $x_1$ and replace all occurrences of the form $x_1 \geq l'$ with $\langle a : C, l' \rangle$, while replace all occurrences the form $x_1 \leq l'$ with $\langle a : \texttt{nnf}(\neg C), 1 - l' \rangle$.

**R**$\bar{m}$. The case $\langle a : \neg m(C), l \rangle \in S_i$ is similar to rule **R**$m$, where we use the bMIP representation of $\bar{g}(m)$ in place of $g(m)$.

**R**$d$. If $\langle c : d, l \rangle \in S_i$ then $S_{i+1} = S_i \cup \gamma(c : d, l)$, where the set $\gamma(c : d, l)$ is obtained from the bMIP representation of $g(d)$ by replacing all occurrences of $x_2$ with $l$ and $x_1$ with $x_c$.

**R**$\bar{d}$. The case $\langle c : \neg d, l \rangle \in S_i$ is similar to rule **R**$d$, where we use the bMIP representation of $\bar{g}(d)$ in place of $g(d)$.

Note that an unique branch is generated in the tableaux of $S_0$, though it can be of exponential length. [5] Furthermore, let us comment the **R**$\sqcup$ rule. By reasoning by case, for $y = 0$, we have $x_1 = 0, x_2 \leq 1, x_2 = l$, while for $y = 1$, we have $x_2 = 0, x_1 \leq 1, x_1 = l$. Therefore, the control variable $y$ simulates the two branchings of the disjunction. A similar argument applies to the other rules.

We say that a constraint set $S'$ obtained from rule applications to $S$ is a *completion* of $S$ iff no more rule can be applied to $S'$. The following can be shown.

**Proposition 1** *Let $S$ be a constraint set. The rules are satisfiability preserving and a completion of $S$ is obtained after a finite number of rule applications.*

**Proposition 2** *Consider $\mathcal{K}\langle \mathcal{T}, \mathcal{A} \rangle$ and let $\alpha$ be a concept assertion axiom $a : C$ or a terminological axiom* $A \sqsubseteq B$. Then in finite time we can determine $glb(\mathcal{K}, \alpha)$ as the minimal value of $x$ such that the completion of $S = \mathcal{T} \cup \mathcal{A} \cup \{\langle \alpha', 1-x \rangle\}$ is satisfiable, where (i) $\alpha' = a{:}\neg C$ if $\alpha = a{:}C$, (ii) $\alpha' = a{:}A \sqcap \neg B$ if $\alpha = A \sqsubseteq B$.

---

[5]The exponential space is due to a well known problem inherited from the crisp case. Indeed, a completion of $S = \{\langle x : C, 1 \rangle\}$ contains at least $2^n + 1$ variables, where $C$ is the concept $(\exists R.d_{11}) \sqcap (\exists R.d_{12}) \sqcap \forall R.((\exists R.d_{21}) \sqcap (\exists R.d_{22}) \sqcap \forall R.((\exists R.d_{31}) \sqcap (\exists R.d_{32}) \ldots \sqcap \forall R.((\exists R.d_{n1}) \sqcap (\exists R.d_{n2})) \ldots)$.

**Example 3** *Let us consider a simplified version of Example 2, by showing that $\mathcal{K} \models \langle \text{Minor} \sqsubseteq \text{YoungPerson}, 0.6 \rangle$ holds, where $\text{Minor} = \leq_{18}$ and $\text{YoungPerson} = \text{Young}$, and that this is the best degree bound. We use $\text{M}, \text{Y}$ and $\text{YP}$ as a shorthand for $\text{Minor}, \text{YoungPerson}$ and $\text{Young}$, respectively. For ease, a variable $x_\alpha$, where $\alpha$ is an assertion is simply written as $\alpha$. We have to consider*

$$S_0 \cup \{\langle b{:}\text{M} \sqcap \neg \text{YP}, 1-x \rangle\} \;,$$

*where $b$ is a new abstract individual. That is, we have to minimize $x$ such that*

$$S_1 = \mathcal{T} \cup \{\langle b{:}\leq_{18} \sqcap \neg \text{Y}, 1-x \rangle, x \in [0,1]\}$$

*is satisfiable. By application of the $\mathbf{R}\sqcap$ rule we get*

$$S_2 = S_1 \cup \{\langle b{:}\leq_{18}, 1-x \rangle, \langle b{:}\neg \text{Y}, 1-x \rangle\} \;.$$

*By abuse of notation, we write $\langle b{:}\neg \text{Y}, 1-x \rangle$ as $b{:}\text{Y} \leq x$. Now, for $x=1$, $S_2$ is satisfiable, while for $x=0$, from $\langle b{:}\leq_{18}, 1 \rangle$, $0 \leq x_b \leq 18$ follows and from $b{:}\text{Y} \leq 0$, $x_b \geq 30$ is required and, thus, $S_2$ is not satisfiable (for $x=0$). For $0 < x < 1$, $0 \leq x_c \leq 18$ should hold. Furthermore, over $[0,30]$ it can be shown that*

$$\bar{g}(\text{Y}) = \{\langle x_1, x_2 \rangle {:} x_1 \leq 10 + 20y, x_2 \geq (1-y), x_1 \geq 10y,\\ x_1 \leq 30, x_1 + 20 x_2 \geq 30y, x_i \in [0,1], y \in \{0,1\}\}$$

*holds (see Equation 3 in the appendix). This means that, from $S_2$, by applying the $\mathbf{R}\bar{d}$ rule to $b{:}\text{Y} \leq x$, we get the set $S_3 = S_2 \cup \{x_b \leq 10 + 20y, x \geq (1-y), x_b \geq 10y, x_b \leq 30, x_b + 20x \geq 30y, y \in \{0,1\}\}$. For $y=0$, $x_b \leq 10$ and $x=1$ have to hold and $S_3$ is still satisfiable. On the other hand, for $y=1$, $x_b \geq 10$ and $x_b + 20x \geq 30$ hold. That is, $x \geq (30 - x_b)/20$. As $10 \leq x_b \leq 18$, the minimal value of $x$ satisfying $S_3$ under this condition is, thus, $x = 3/5$. Therefore, the minimal solution $x$ satisfying $S_3$ is $x = 3/5$.*

## 4  RELATED WORK

The first work on fuzzy DLs is due to Yen ([19]) who considered a sub-language of $\mathcal{ALC}$, $\mathcal{FL}^-$ [2]. However, it already informally talks about the use of modifiers and concrete domains. Though, the unique reasoning facility, the subsumption test, is a crisp yes/no question. Tresp ([18]) considered fuzzy $\mathcal{ALC}$ extended with a special form of modifiers, which are a combination of two linear functions, as we described in the appendix. min, max and $1-x$ membership functions has been considered and a sound and complete reasoning algorithm testing the subsumption relationship has been presented. Similar to our approach, a linear programming oracle is needed. Assertional reasoning has been considered by Straccia ([15]), where fuzzy assertion axioms have been allowed in fuzzy $\mathcal{ALC}$ (with min, max and $1-x$ functions), concept modifiers are not allowed however. He also introduced the BDB problem and provided a sound and complete reasoning algorithm based on completion rules ([16] provides a translation of fuzzy $\mathcal{ALC}$ into classical $\mathcal{ALC}$). For an application see [10]. In the same spirit [6] extend Straccia's fuzzy $\mathcal{ALC}$ with concept modifiers of the form $f_m(x) = x^\beta$, where $\beta > 0$. A sound and complete reasoning algorithm for the graded subsumption problem, based on completion rules, is presented. [13] starts addressing the issue of alternative semantics of quantifiers in fuzzy $\mathcal{ALC}$ (without the assertional component). No reasoning algorithm is given. Finally, [5] considers $\mathcal{ALC}$ under arbitrary t-norm and reports, among others, a procedure deciding $\models \langle C \sqsubseteq D \geq 1 \rangle$ and deciding whether $\langle C \sqsubseteq D \geq 1 \rangle$ is satisfiable, by a reduction to the propositional BL logic.

## 5  CONCLUSIONS AND OUTLOOK

We have presented fuzzy $\mathcal{ALC}(\mathbf{D})$ showing that its representation and reasoning capabilities go clearly beyond current approaches to fuzzy DLs. We believe that the fuzzy extension of $\mathcal{ALC}(\mathbf{D})$ allows to express naturally a wide range of concepts of actual domains, for which a classical representation is unsatisfactory. Fuzzy $\mathcal{ALC}(\mathbf{D})$ enhances current approaches as we allow arbitrary bMIP-representable concept constructors, modifiers and fuzzy domain predicates to appear in a $\mathcal{ALC}(\mathbf{D})$ knowledge base. The entailment and the subsumption relationship hold to a certain degree. We also presented a solution to the BDB problem based on a minimization problem on bMIP.

Future work involves the extension of fuzzy $\mathcal{ALC}(\mathbf{D})$ to $\mathcal{SHOIN}(\mathbf{D})$, the theoretical counterpart of OWL DL. Another direction is in extending fuzzy DLs with *fuzzy quantifiers*, where $\forall$ and $\exists$ are replaced with fuzzy quantifiers like most, some, usually and the like (see [13] for a preliminary work in this direction). This allows to define concepts like

```
TopCust = Customer ⊓ (Usually)buys.ExpensItem
ExpensItem = Item ⊓ ∃price.High .
```

Here, the fuzzy quantifier Usually replaces the classical quantifier $\forall$ and High is a fuzzy concrete predicate. Fuzzy quantifiers can be applied to inclusion axioms as well, allowing to express, e.g.

(Most)Bird ⊑ FlyingObject .

Here the fuzzy quantifier Most replaces the classical universal quantifier ∀ assumed in the inclusion axioms. The above axiom allows to state that most birds fly.

## A  ON MEMBERSHIP FUNCTIONS

As a building blocks for membership function specification, we consider linear functions and the combination of two linear functions: let $[k_1, k_2]$ be an interval in $\mathbb{Q}$, $L: [k_1, k_2] \to [0, 1]$ is defined as

$$L_{[k_1,k_2]}(x; f_1, c, f_2) = \begin{cases} f_1(x) & \text{if } k_1 \leq x \leq c \\ f_2(x) & \text{if } c \leq x \leq k_2 \end{cases}$$

where $c \in [k_1, k_2]$, $f_1$ and $f_2$ are linear functions $f_i: [k_1, k_2] \to [0, 1]$, $f_i(x) = m_i x + q_i$, $m_i, q_i \in \mathbb{Q}$, such that $f_1(c) = f_2(c) \geq 0$. Notice that for modifiers, we require that the domain is $[0, 1]$. Furthermore, note that the modifiers in [18] are a special case as additionally $f_1(c) = f_2(c)$, $m_1 > 0$ and $m_2 < 0$ should hold. As an application of linear combination functions, we may define, e.g. the modifier very as $L_{[0,1]}(x; \frac{2}{3}x, 0.75, 2x - 1)$. While the modifier $m(x) = x^2$ ([6]) cannot be bMIP-represented, the previous definition may be seen as an approximation of it. Multiple combinations of linear functions may be used to represent the membership function depicted in Figure 1.

For the sake of concrete illustration, we first show how to represent the combination of two linear functions as a bMIP. It will be then evident that any combination of more than two linear functions can be obtained in a similar way and, thus, the trapezoidal functions are just a special case. So, consider $L_{[k_1,k_2]}(x; f_1, c, f_2)$. There are several cases to consider according to the value of $m_i$ ($< 0, > 0$ and 0). In order to represent $L$ as a bMIB, we have to define the graph $g(L) = \{\langle x_1, x_2 \rangle : L(x_1) \geq x_2\}$ as the solutions of a bMIP. However, as we may have negation on front of modifiers and fuzzy domain predicates, $\bar{g}(m) = \{\langle x_1, x_2 \rangle : L(x_1) \leq x_2\}$ should be bMIP-representable as well. We just consider the former case as the latter can be developed in a similar way. We have that $f_1(k_1) \geq 0$ and $f_2(k_2) \geq 0$. Under this condition, $g(L)$ can be split into two sets $X_1$ and $X_2$, $g(L) = X_1 \cup X_2$, where $X_1 = \{\langle x_1, x_2 \rangle : f_1(x_1) \geq x_2, k_1 \leq x_1 \leq c, 0 \leq x_2 \leq 1\}$, while $X_2 = \{\langle x_1, x_2 \rangle : f_2(x_1) \geq x_2, c \leq x_1 \leq k_2, 0 \leq x_2 \leq 1\}$. From the $X_i$, we can build matrixes $A_i^j$ and rational positive vectors $\mathbf{b}_i^j$ ($i, j = 1, 2$) such that $X_i$ can be written as the set $X_i = \{\mathbf{x} : A_i^1 \mathbf{x} \geq \mathbf{b}_i^1, A_i^2 \mathbf{x} \leq \mathbf{b}_i^2\}$. Now we introduce a 0-1 valued control variable $y$ in order to merge the two sets $X_1$ and $X_2$ into a bMIP. Indeed, we define for vectors $\mathbf{w}_i^j$ of rational values $X_{12} = \{\mathbf{x} : A_1^1 \mathbf{x} \geq (1-y) \cdot \mathbf{b}_1^1 + y \cdot \mathbf{w}_1^1, A_1^2 \mathbf{x} \leq (1-y) \cdot \mathbf{b}_1^2 + y \cdot \mathbf{w}_1^2, A_2^1 \mathbf{x} \geq y \cdot \mathbf{b}_2^1 + (1-y) \cdot \mathbf{w}_2^1, A_2^2 \mathbf{x} \leq y \cdot \mathbf{b}_2^2 + (1-y) \cdot \mathbf{w}_4^2\}$, Then, it can be verified that there is a suitable choice of $\mathbf{w}_i^j$ such that for $y = 0$, $X_{12} = X_1$, while for $y = 1$ $X_{12} = X_2$ and, thus, $X_{12} = g(L)$ and from $X_{12}$ a bMIP can easily be obtained. The graph $\bar{g}(L)$ can then be defined in a similar way. For instance, Young, restricted to $[0, 30]$, can be defined as $L_{[0,30]}(x; 1, 10, (30-x)/20)$ and, thus, it can be shown that $\bar{g}(L)$ is

$$X_{12} = \{\langle x_1, x_2 \rangle : x_1 \leq 10(1-y) + 30y, x_2 \geq (1-y), \\ x_1 \geq 10y, x_1 \leq 30y + 30(1-y), x_1 + 20x_2 \geq 30y\} . \tag{3}$$

This completes the first part. Now, in order to extend Young to range over, say, $[0, 200]$ and not just over $[0, 30]$ (recall that $\text{Young}(x) = 0$ for $x \geq 30$) we have to reapply the above procedure again to the sets $X_{12}$ and $X_3$, where $X_3 = \{\langle x_1, x_2 \rangle : x_1 \geq 30, x_2 = 0\}$ (this will introduce another control variable $y_1$), obtaining the set $X_{123}$. Therefore, Young is bMIB representable with two control variables. In general, it can be verified that the above procedure can iteratively be applied to the union of $n \geq 2$ sets of the form $X_i$, by means of the introduction of $n - 1$ control variables. In particular, trapezoidal functions can be represented as bMIP using at most four control variables ($n = 5$).

The attentive reader will notice that a difficulty arises in representing crisp sets, such as e.g. $\leq_{18}$, as they present a discontinuity. To overcome partially to this situation, we may rely on a linear combination of the form $L_{[0,18+\epsilon]}(x; 1, 18, (18+\epsilon - x)/\epsilon)$ for a sufficiently small $\epsilon > 0$ and then extend it to range over, say $[0, 150]$, by combining the previous function with $f(x) = 0$, for $18 + \epsilon \leq x \leq 150$, in a similarly way as we did for Young (so, two control variables are needed).

## B  RULES FOR LUKASIEWICZ LOGIC

**R**$A$.  If $\langle \alpha, l \rangle \in S_i$ and $\alpha$ is an atomic assertion of the form $a: A$ or $(a, b): R$ then $S_{i+1} = S_i \cup \{x_\alpha \geq l\}$.

**R**$\bar{A}$.  If $\langle a: \neg A, l \rangle \in S_i$ then $S_{i+1} = S_i \cup \{x_{a: A} \leq 1 - l\}$.

**R**⊓.  If $\langle a: C \sqcap D, l \rangle \in S_i$ then $S_{i+1} = S_i \cup \{\langle a: C, x_1 \rangle, \langle a: D, x_2 \rangle, y \leq 1 - l, x_i \leq 1 - y, x_1 + x_2 = l + 1 - y, x_i \in [0, 1], y \in \{0, 1\}\}$, where $x_i$ is a new variable, $y$ is a new control variable.

**R**⊔.  If $\langle a: C \sqcup D, l \rangle \in S_i$ then $S_{i+1} = S_i \cup \{\langle a: C, x_1 \rangle, \langle a: D, x_2 \rangle, x_1 + x_2 = l, x_i \in [0, 1]\}$, where $x_i$ is a new variable.

**R**∃.  If $\langle a: \exists R.C, l \rangle \in S_i$ then $S_{i+1} = S_i \cup \{\langle (a, b): R, x_1 \rangle, \langle b: C, x_2 \rangle, y \leq 1 - l, x_i \leq 1 - y, x_1 + x_2 = l + 1 - y, x_i \in [0, 1], y \in \{0, 1\}\}$, where $x_i$ is a new variable, $y$ is a new control variable and $b$ is a new abstract individual. The case for concrete roles is similar.

**R∀.** If $\{\langle a{:}\forall R.C, l_1\rangle, \langle (a,b){:}R, l_2\rangle\} \subseteq S_i$ then $S_{i+1} = S_i \cup \{\langle a{:}C, x\rangle, x \geq l_1 + l_2 + 1, x \leq y, l_1 + l_2 - 1 \leq y, l_1 + l_2 \geq y, x \in [0,1], y \in \{0,1\}\}$, where $x$ is a new variable and $y$ is a new control variable. The case for concrete roles is similar.

**R$m$.** If $\langle a{:}m(C), l\rangle \in S_i$ then $S_{i+1} = S_i \cup \gamma(a{:}C, l)$, where the set $\gamma(a{:}C, l)$ is obtained from the bMIP representation (see appendix) of $g(m)$ as follows: replace in $g(m)$ all occurrences of $x_2$ with $l$. Then resolve for $x_1$ and replace all occurrences of the form $x_1 \geq l'$ with $\langle a{:}C, l'\rangle$, while replace all occurrences the form $x_1 \leq l'$ with $\langle a{:}\mathtt{nnf}(\neg C), 1-l'\rangle$.

**R$\bar{m}$.** The case $\langle a{:}\neg m(C), l\rangle \in S_i$ is similar to rule **R$m$**, where we use the bMIP representation of $\bar{g}(m)$ in place of $g(m)$.

**R$d$.** If $\langle c{:}d, l\rangle \in S_i$ then $S_{i+1} = S_i \cup \gamma(c{:}d, l)$, where the set $\gamma(c{:}d, l)$ is obtained from the bMIP representation of $g(d)$ by replacing all occurrences of $x_2$ with $l$ and $x_1$ with $x_c$.

**R$\bar{d}$.** The case $\langle c{:}\neg d, l\rangle \in S_i$ is similar to rule **R$d$**, where we use the bMIP representation of $\bar{g}(d)$ in place of $g(d)$.

Let us comment the **R⊓** rule. By reasoning by case, for $y = 0$, we have $x_i \leq 1, x_1 + x_2 = l+1$, while for $y = 1$, we have $l = 0, x_i = 0$. These two cases correspond to $\max(0, x_1 + x_2 - 1) \geq l$, which is true if $l = 0$ ($y = 1$) or $x_1 + x_2 - 1 \geq l$ ($y = 0$) with $x_1 + x_2 - 1 \geq 0$. Therefore, the control variable $y$ simulates the two alternatives of the max operator in the definition of conjunction. A similar argument applies to the other rules.